\documentclass[sigconf]{acmart}

\AtBeginDocument{%
  \providecommand\BibTeX{{%
    \normalfont B\kern-0.5em{\scshape i\kern-0.25em b}\kern-0.8em\TeX}}}

\copyrightyear{2021}
\acmYear{2021}
\setcopyright{acmlicensed}\acmConference[AIES '21]{Proceedings of the 2021 AAAI/ACM Conference on AI, Ethics, and Society}{May 19--21, 2021}{Virtual Event, USA}
\acmBooktitle{Proceedings of the 2021 AAAI/ACM Conference on AI, Ethics, and Society (AIES '21), May 19--21, 2021, Virtual Event, USA}
\acmPrice{15.00}
\acmDOI{10.1145/3461702.3462619}
\acmISBN{978-1-4503-8473-5/21/05}

\begin{document}
\fancyhead{} 
\title{AI and Shared Prosperity}


\author{Katya Klinova}
\affiliation{%
  \institution{Partnership on AI}
  \streetaddress{Street address}
  \city{San Francisco, California}
  \country{USA}}
\email{katya@partnershiponai.org}

\author{Anton Korinek}
\affiliation{%
  \institution{University of Virginia}
  \streetaddress{Street address}
  \city{Charlottesville, Virginia}
  \country{USA}}
\email{akorinek@virginia.edu}


\begin{abstract}
Future advances in AI that automate away human labor may have stark implications for labor markets and inequality. This paper proposes a framework to analyze the effects of specific types of AI systems on the labor market, based on how much labor demand they will create versus displace, while taking into account that productivity gains also make society wealthier and thereby contribute to additional labor demand. This analysis enables ethically-minded companies creating or deploying AI systems as well as researchers and policymakers to take into account the effects of their actions on labor markets and inequality, and therefore to steer progress in AI in a direction that advances shared prosperity and an inclusive economic future for all of humanity.
\end{abstract}



\begin{teaserfigure}
    \centering
    \includegraphics[width=\textwidth]{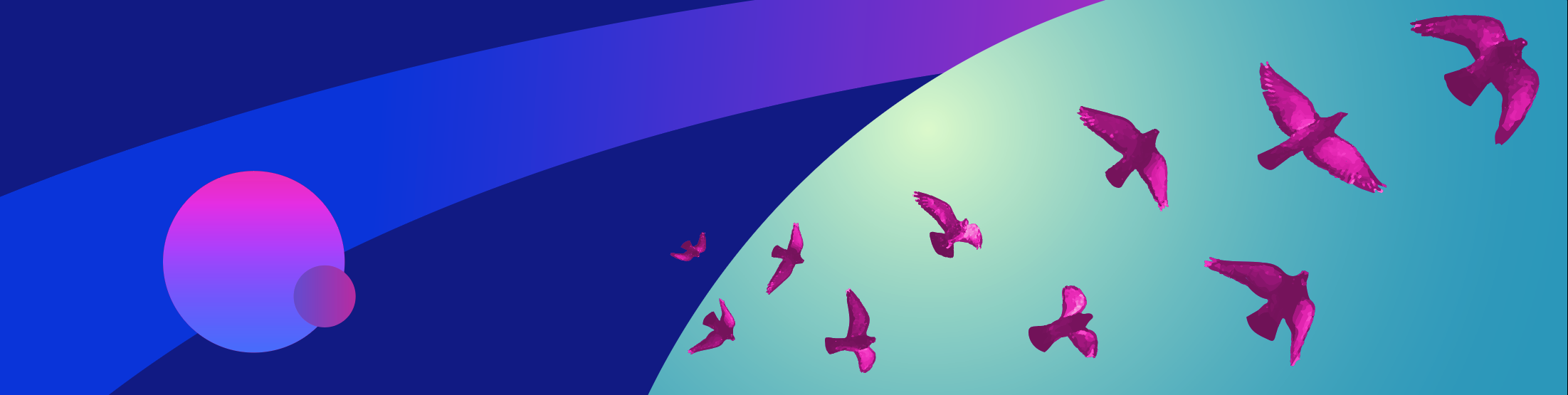}
	\medskip
\end{teaserfigure}

\maketitle

\nocite{pai_ra}
\nocite{korineks18}

\section{Introduction}
One of the major avenues through which the advancement of Artificial Intelligence (AI) is affecting society and its people is by redistributing economic opportunities and earning prospects. Such redistribution can be equitable and inclusive, or biased in favor of certain groups that get to benefit from economic power concentrating in their hands \cite{korineks19}. 

Up until recently, many technologists believed that technological progress always “lifts all boats'' and automatically leads to shared prosperity once the economy has gone through an adjustment period. This has left many to ignore the consequences of their inventions for economic inequality. However, since the 1980s, technological progress has been accompanied by a significant increase in inequality. For example, \cite{autor2015there} surveys a significant body of economic literature that finds that automation in the US has led to a polarization of the labor market, whereby middle-income jobs that used to perform routine tasks were replaced by lower-income jobs, while those at the top of the income distribution experienced significant gains, leading to an increase in economic inequality. \cite{schwellnus2017decoupling} document a similar phenomenon across the OECD, a club encompassing the richest countries of the world, over the past two decades: that median wages - the compensation of a typical worker - grew at a slower pace than overall productivity, and that the gap between the two has been increasing. 

In recent years it has - fortunately - become common to see demands that “AI should be human-centered, with transparency and accountability as paramount features” (NextGen report, 2020). However, those are often immediately - and unfortunately - followed by statements like “Overwhelmingly, AI will disrupt labor markets and the economy around the world,” which are taken as immutable facts. There seems to be a big difference in attitude towards AI violating notions of fairness, accountability and transparency, versus the prospect of AI disrupting the economic order, increasing inequality, and endangering the livelihood of millions of workers. The former is justifiably seen as unacceptable, while the latter is often taken as a given and treated as an unavoidable consequence of AI advancement. 

The difference in attitude seems in part driven by the ethical foundations used to evaluate advances in AI. Under traditional deontological foundations, developing an AI system that operates and performs all its actions while complying with all of society’s traditional ethical rules is considered fair game, even if the system has the side effect of triggering general equilibrium effects in the economy that lead to the displacement of millions of jobs. By contrast, under more consequentialist foundations, it is more natural that such massive job displacement would be considered a violation of ethical norms.

The distinction between the two approaches also has significant implications for how society is expected to respond to job disruption: if it is considered fair game for the developers and deployers of AI to disrupt labor markets and impose losses on millions of workers, then the burden of adjustment is on workers: it is their duty to face their losses, and they must continuously upskill in order to remain relevant in the labor market that is being reshaped by AI advancement. The implication for governments is that they are well-advised to strengthen social safety nets, expand training programs and prepare for benefits like a Universal Basic Income, all financed by taxpayers, not the perpetrators of the disruptions. In other words, it is society and its people, not the AI developers, who are expected to bear the burden of ensuring that AI advancement does not cut people off their sources of income. 

By contrast, if it is considered unethical for the developers and deployers of AI to disrupt labor markets, then the burden is on the AI industry to consider the impact of the technologies they are shaping on labor markets and employment opportunities. And it would be incumbent upon the AI industry to guide its decisions around the development of AI to avoid massive elimination of jobs and make the adjustments to a new technological and economic reality less burdensome and costly for workers and taxpayers.

This paper builds on the recent economic literature discussing the impact of automation technology on labor demand and lays out a framework for systematically evaluating and predicting the impact of new AI-based technologies on labor income. We propose a step-by-step procedure that can be used by interested stakeholders in the AI industry and research community without prior knowledge of economics to evaluate which AI applications are likely to “lift all boats” and support shared prosperity, and which ones are likely to contribute to greater inequality. Evaluating this question throughout the AI development process is a critical step towards assuring that AI advancement is ethical and creates a fair and inclusive economic future for humanity \cite{korinek2020integrating}. We hope that such an evaluation will inform the choices of researchers, developers, innovators, investors and consumers in the field of AI, and that it will help to practically execute the idea of steering the direction of AI progress to benefit the workers instead of displacing them \cite{acemoglu2020wrong,korinek2020steering}.

A decrease in labor demand can manifest itself as reduction in wages, employment, or both. Hence gauging the change in labor demand allows to understand the direction and magnitude of impact on workers’ incomes and financial well-being. We should note that assessing AI’s impact on labor demand does not necessarily capture the impacts that technology has on other aspects of well-being and overall job quality, like safety, level of physical strain, schedule predictability, freedom from surveillance, etc. However, job quality is frequently correlated with wages levels as higher labor demand also gives workers more bargaining power to ask for better job conditions.

The rest of this paper is structured as follows: Section 2 discusses why the current trajectory of AI advancement risks exacerbating economic inequality; Section 3 presents ways to practically incorporate the consideration of economic inequality into the AI research and development process and proposes a high-level framework for directing AI progress for Shared Prosperity; Section 4 makes a case for AI researchers and developers to recognize the responsibility of ensuring that their creations support economic inclusion; Section 5 outlines the framework’s limitations and open questions for future work; Section 6 concludes.

\section{Advances in AI and Employment}

Traditional economic theory views capital and labor as the main \emph{factors of production} in the economy. Capital includes machines, equipment, etc. The factor labor can be distinguished by geography or by levels of education, e.g.\ into lower-skilled, medium-skilled and higher-skilled labor. Depending on the application, it may also be useful to distinguish workers along specific occupations that may be differentially affected by an innovation.

Technological change may boost the returns on some or all of these factor owners. But some types of technological change lead to starkly diverging impacts on different factor owners \cite{korineks19}. For example, they may increase the returns on capital but not labor, or of higher-skilled workers but not lower-skilled workers, or -- to offer a very specific example -- of AI engineers but not radiologists. By implication, they may shake up who in society has access to gainful employment opportunities and for whom those opportunities become harder to obtain as a consequence of falling demand for their skills.

AI may change workers' access to economic opportunity even more starkly than the redistributions generated by past waves of technological progress.  Let us provide a clear example of the type of redistributions that AI may generate: an e-commerce business that scales up might displace a large number of local mom-and-pop stores, resulting in a concentration of gains from selling goods to consumers. To provide another example, a shift towards consuming news through social media may disrupt traditional patterns of news consumption and may cause local newspapers to lose advertising revenues, leading to large job losses at local newspapers and, again, resulting in a concentration of gains in the hands of a few large companies. AI may also generate significant redistributions across countries \cite{korinek2021artificial}.

In the following, let us discuss three common types of technological change that are frequently discussed and that may have large effects on workers' earnings opportunities: automation, skill-biased technological change, and human augmentation.

\subsubsection*{Automation}

Automation displaces human labor with machines. Frequently, higher-skilled workers are employed to design systems that automate away the jobs of lower-skilled workers. A cursory look might suggest that automating a job always has negative effects on workers while creating jobs always has positive effects on workers. But it is more complicated than that. Automation of human labor is not, in itself, undesirable. The history of progress since the Industrial Revolution is a story of relentless automation that has contributed to rising living standards. Moreover, there are many examples when automation allowed us to make work less dangerous and less physically taxing. 

If the automation of human tasks is accompanied by the creation of new tasks for humans, then the adverse effects on labor demand are offset \cite{acemoglu2019automation}. Empirical evidence from the US suggests that while tasks automated and reinstated by technological change used to balance out during the four decades following WWII, the past three decades have seen task displacement significantly outpacing reinstatement. AI is expected to continue this trend and may even accelerate it \cite{acemoglu2020wrong}.

Moreover, if the new tasks that AI advancement creates require a much higher level of skills or educational attainment compared to the tasks being displaced, these new tasks will bring little to no economic relief to the workers whose jobs get automated, even if the new jobs match or exceed the displaced jobs in volume.

\subsubsection*{Skill-biased technological change}

A type of technological change that disproportionately benefits those with comparatively high levels of educational attainment is referred to as skill-biased. Skill-biased technological change is poised to exacerbate society’s structural inequalities, especially in countries with low economic mobility.

Technological change does not have to be skill-biased. In fact, the first wave of the Industrial Revolution is generally viewed as having been biased in favor of unskilled workers, who suddenly had much greater earning opportunities. Likewise, there is nothing inherent about AI that makes its applications necessarily be skill-biased, or labor-saving. AI can be used to complement lower-skilled workers, making their labor valued more highly. It could also be used to expand the economic possibilities of people who previously had limited access to training and education, or for whom education was prohibitively costly. In other words, AI advancement could be economically inclusive, but making it so requires deliberate action.

\subsubsection*{Human Augmentation}


It is commonly suggested that developing AI systems that augment human workers instead of displacing them would be a good recipe to ensure that advances in AI benefit workers. Some AI firms have even started to use this language when describing their products in promotional materials and sales pitches. While “human-augmenting AI” as a generic goal seems more desirable than measuring the progress of AI by how well it automates away humans, it is important to note that human- or labor-augmenting AI can still result in the displacement of workers and reductions in wages, because labor-augmenting AI may still be labor-saving -- it all depends on whether the firm will cut its prices and how much demand will respond to the price cut.

For a simple example, suppose that a firm employs workers to produce a product that is sold to consumers. Suppose further that the firm develops and deploys labor-augmenting AI technology that allows each worker to increase their output per hour by 10\%. With this technology, 10\% fewer workers are needed to produce the same amount of output. If the firm cannot increase its production volume or the number of units it can sell per year, it may have to make redundant those 10\% of its workers.
By contrast, if the firm lowers the price of its product as a result of its reduced labor costs, and if consumers are willing to buy a lot more of the product at the lower price, then the firm might keep or even grow the size of its original workforce to satisfy the increase in the demand for its product. Whether this happens or not depends on how responsive consumer demand for the product is to the changes in its price (in economic language, it depends on whether the elasticity of product demand is greater or less than 1). In a competitive market, the firm is indeed likely to lower its price when its costs decline. And even in a monopolistic market the firm might lower its price, for example if it aims to keep its markup constant.

This stylized example illustrates that developing AI that is “augmenting humans” is not enough to ensure that employment and wages are not reduced. Because labor-augmenting technology allows firms to employ fewer workers to produce the same amount of goods, something needs to compensate for the reduction in labor demand that such technology will create in order to avoid job losses.
\medskip

These three examples illustrate that the effects of new technologies on labor markets are not always straightforward to assess, and that a systematic economic framework is needed to sort through the different effects. In the following section, we propose the outlines of such a framework, allowing researchers, companies, and policymakers to evaluate the overall impact of a new technology on labor markets.

As we discuss the effects of technology on wages and inequality, we do not take a stance on what particular social welfare function to embrace, but our results on what redistributions a technological innovation will give rise to are relevant for any inequality-averse social welfare function.

\section{Framework for Evaluating AI’s Impact on Labor Demand}

This section describes a set of questions that represent a heuristic to systematically evaluate the overall, or general equilibrium, impact of an AI application on labor demand and, consequently, on employment and wages. We will use customer service chatbots and autonomous grocery delivery vehicles as illustrative examples when describing the logic of the framework.

Customer service chatbots are applications frequently powered by Natural Language Processing machine learning models that simulate the behavior of human customer support agents. For the purposes of this discussion, we will refer to applications that respond textually or vocally to textual or vocal customer service requests as chatbots. In today's world, human customers frequently encounter such chatbots, for example when texting or calling customer support number, responding to a “How can I help you today?” automatic pop-up prompts on commercial websites and in other contexts.

Autonomous delivery vehicles are driverless vehicles with a space to transport goods, but not people. At the time of writing, autonomous delivery is not widely commercially available, but a few startup companies are piloting them in a limited set of locations.\footnote{See, for example: https://venturebeat.com/2021/01/27/starship-raises-17-million-to-send-autonomous-delivery-robots-to-new-campuses/ and https://medium.com/nuro/california-dmv-grants-nuro-first-ever-av-deployment-permit-ca424ebd2} For the purposes of this discussion, we will examine autonomous grocery delivery, while noting that autonomous delivery vehicles could be used to deliver other types of goods as well.

The framework that we propose in the following does not provide precise quantitative estimates of the magnitude of effects. Often mapping relative magnitudes and the directions of effects is in itself already very instructive and permits to understand the direction of the overall impact on labor demand. However, we hope that the proposed framework will inspire follow-up work on how to flesh out the described effects in more detail and at a more quantitative level \cite{pai_ra}. Depending on the particular AI application, we also note that some steps may be skipped if the magnitude of the described effects is not deemed significant enough to merit a deep investigation.

\subsection{Direct Effects}

The direct effects reflect the workers who are directly hired or displaced because of the introduction of a new AI application in a company, while holding everything else fixed. We capture these direct effects via the following two questions: 

\medskip

\noindent \textit{(1) Which types of workers will be displaced by the introduction of a new AI system, and in which geographies are they located?}

It is important to capture the skill level and geographic location of those workers in order to understand if any of the compensatory effects below will be of relevance to them. For example, if an AI application displaces workers without a college degree in one set of locations and creates an equal number of jobs requiring advanced degrees in another location, the displaced workers will not be able to compete for the newly created jobs.

In the chatbots example, the workers directly displaced are customer service associates located around the world, with major centers in the United States, India, and the Philippines. Those are predominantly formal sector jobs with predictable schedules and earnings. Skill requirements vary.

The workers directly displaced by autonomous grocery delivery vehicles are delivery associates often employed as independent contractors, also referred to as gig workers. Their jobs are frequently precarious and lacking earnings predictability and benefits, but the work can be physically demanding and prone to injury. There usually are no degree requirements associated with delivery jobs. 
Notably, autonomous delivery vehicles would also “displace” unpaid work by households who shop for and bring their groceries home by themselves, without calling a delivery person. 

\medskip

\noindent\textit{(2) For what types of workers will new demand be created by the introduction of the AI application, and in which locations?}

The introduction of chatbots will create demand for software engineers customizing and maintaining the bots, chatbot platform sales and customer success people, product and marketing managers. Most of those jobs will require a college or advanced degree and will likely be geographically concentrated. They will also likely be fewer in number compared to the number of displaced customer service associates identified in step 1, because one chatbot development company can service many corporate clients.

Introduction of autonomous grocery delivery vehicles will create jobs similar to the above at companies producing those vehicles, but in addition would create a need for workers assembling grocery orders and loading them onto the autonomous delivery vehicles (``pickers''). Those jobs will be similar in quality and skill requirements to the jobs of displaced delivery gig workers but might be safer and higher in volume if a significant number of households switch from shopping for and delivering their own groceries to using an autonomous delivery vehicle-powered service.

\subsection{Demand Effects}

\textit{(3) Will the innovation increase demand for the company's products because it will lower its prices or increases the quality of its products? Will the increase in product demand translate into higher demand for workers? For what categories of workers and in which geographies?}

A company that cuts its costs by replacing some or all of its human customer support agents with an automated chatbot might lower the costs of its main product because of the lower expenses on customer support. Moreover, it might offer more customer support services and choose to provide it in contexts where it was not used before (for example for marketing, with a proactive “Can I help you…?” pop-up when a prospective customer visits the company’s webpage). This might not have much impact on how many human customer support agents it employs, but it may somewhat raise labor demand for all types of workers across the company.

If grocery deliveries are made easier and less costly by the introduction of autonomous delivery, households will demand more grocery deliveries, increasing the revenue and associated employment of pickers at grocery stores. However, there are likely few effects on the overall volume of groceries purchased since grocery consumption is quite inelastic and quickly reaches a point of saturation.

\subsection{Vertical Effects}

\textit{(4) How does additional product demand affect labor demand along the supply chain of the innovating company?} 

In the chatbot example, the innovating company will need fewer workstations for customer support, less office space, and fewer office services, resulting in lower labor demand among those suppliers. However, if the company can expand the demand for its products because of lower prices or better marketing, then it will also raise the demand for its intermediate inputs, which may add labor demand to the economy.

For the grocery delivery vehicles, we do not anticipate significant vertical effects since total demand for groceries will not change significantly.

\subsection{Horizontal Effects}

The horizontal effects capture how companies that produce goods and services that are substitutes or complements to the innovating company will be affected. Economists refer to goods or services as substitutes if they can easily be used for the same purpose, e.g.\ taxi rides or Uber rides. Goods or services are complements when an increase in the consumption of one makes it more desirable to consume the other, e.g.\ coffee and cream.

\medskip

\noindent\textit{(5) How will competing companies producing substitute products/services be affected? Will they need to downsize their workforces? In which geographies and for which categories of workers?}

For the company introducing chatbots, lower prices may undercut its competitors and lead to cuts in their workforce.
In the case of autonomous grocery delivery, services associated with getting to and from the grocery store might be impacted, resulting in a decrease in car sales and in the use of public transportation, impacting the associated employment. 

\medskip

\noindent\textit{(6) What complementary processes might be disrupted, what groups of workers do they employ and where?}

The customer service chatbots are not anticipated to lead to significant effects on complementary goods.

If a significant number of households switch from buying their own groceries to using delivery services as a result of the introduction of autonomous delivery vehicles, it might affect the volumes of purchases of services that rely on the physical presence of customers at grocery stores, for example small vendors in front of the store. The effects are not likely to be major.

\subsection{Factor Reallocation}
\textit{(7) How will wages adjust to reflect the new balance of labor demand and supply resulting from these changes?
}

Our final question considers how equilibrium in the labor market is affected. The workers displaced by technological progress will, after an adjustment period, be redeployed in different companies or sectors, increasing overall output and wealth in the economy. However, when workers need to compete for new jobs, they will push down wages in their sector. And conversely, for workers who are in higher demand, wages will rise. This implies income redistributions that have the potential to exacerbate - or to mitigate - inequality, depending on the specifics of the situation. 

In our example of chatbots, the layoffs of customer service representatives will put downward pressure on the wages of unskilled workers, whereas the hiring of additional AI engineers will put upward pressure on their wages. Moreover, capital owners will benefit from the greater returns that the company earns. Ultimately, the described effects are likely to increase inequality.

For the grocery delivery vehicles, the results will be similar, leading to a redistribution of income from unskilled workers to AI engineers, although the losses of unskilled workers are likely mitigated by the fact that many of the displaced drivers can be redeployed as pickers in grocery stores. 



\medskip

It is important that the field of AI introduces a practice of systematically evaluating these effects to gauge the impact of their inventions on the income prospects of different groups of workers and to ensure a fair distribution of the gains from progress. We view it of particular importance for AI developers to pay attention not to disadvantage those groups of society who already are marginalized and who have limited access to retraining opportunities.

\section{Responsibility of the AI Community}

Progress in AI is unlikely to bring about an inclusive economic future if the direction of AI development is determined solely by market forces, which tend to favor efficiency but do not ensure that the gains from progress are distributed equitably. For example, in recent decades the labor share of income in the US has decreased, and the wages of non-college educated workers have stagnated and have not shared in the productivity gains generated by technological progress \cite{autor2019work}.

Market prices do not always reflect the true social costs and benefits generated by a business activity and therefore may provide misguided incentives. For example, in a free market, prices do not reflect the costs of environmental damage produced by a carbon-emitting enterprise. As a result, the costs are not borne by the polluter but instead by society as a whole. To make the polluting enterprise internalize the full costs of its operations, economic theory suggests taxing activities that produce externalities (in this case, carbon emissions). But even in absence of carbon taxes, companies still have a moral responsibility to take on voluntary commitments to cut their emissions and minimize the un-internalized cost to society from their operations.

Likewise, market prices do not reflect distributive concerns. When companies produce technologies that induce undesirable shifts in labor demand and, consequently, reduce the labor share of national income, there is a cost to society in the form of a more unequal income distribution, disrupted livelihoods, distressed families and communities and sometimes even “deaths of despair” \cite{autor2017disappears}. These are costs that companies do not internalize, akin to environmental externalities. In absence of a regulatory framework to internalize the cost of job-displacing innovation, it is up to developers to behave responsibly in how they handle the redistributive power of AI. 

However, aside from voluntary action on the part of AI developers to consider the economic interests of workers in the AI development process, government can also take steps to ensure that the regulatory environment does not provide incentives for excessive automation and ever greater concentration of earning opportunities. For example, the incentives for innovators are affected by tax policies -- current tax regimes that favor capital over labor \cite{acemoglu2020does} and policies that limit labor mobility \cite{pritchett2020future} create strong incentives to develop AI applications that focus disproprotionately on labor-saving use cases. If these policies remain in place, AI advancement might bring about levels of automation well above what is socially optimal, to the disproportionate detriment of the economically vulnerable workers with limited access to retraining opportunities.

The risk of excessive automation is increasingly recognized by scholars on the “future of work.” But unfortunately, the current discourse places too much of the burden of adjustment to the changing technological landscape on workers and governments, including developing country governments. Questions around the role and responsibility of the AI industry and the AI research community remain relatively neglected.

\section{Limitations and Call for Future Work}
The set of heuristics described in Section 3 is an early attempt at defining the questions that the AI industry and research community ought to be asking in order to evaluate the likely impact of their choices on inequality and availability of gainful employment opportunities for workers, especially the more economically vulnerable workers with lower levels of educational attainment and limited access to retraining opportunities. This attempt has both gaps and limitations. A non-exhaustive list of those is below. 

First, the economic framework we outlined is only the beginning of a research agenda to develop models of the impact of advances in AI on earning opportunities. Many different approaches are possible, and we hope that many will be considered, further developed, and refined by both the AI ethics and the economics community. 

Second, there is often a considerable amount of uncertainty when it comes to estimating the magnitude of the likely impacts of AI advancement on labor demand, especially so for the second- and third-round economic effects, such as effects that propagate internationally through trade, global supply chains or other mechanisms.  AI developers and scientists working on basic research which can subsequently be used to enable many different kinds of applications face an additional layer of uncertainty. The heuristics described above could be improved by introducing weights that reflect the associated levels of uncertainty about each of the effects being considered. But even when the level of uncertainty seems overwhelmingly high, we want to caution against the AI industry and research community excusing themselves from the responsibility to think about the likely impact of their actions and choices on economic inclusion. To return to our analogy to environmental effects, just because climate science does not yet allow us to precisely attribute climate effects to the energy management and waste management choices does not mean that our choices should be reckless. The same holds for the effects of AI on inequality. It is incumbent upon members of the AI community to think about the economic future that they are so powerfully contributing to for societies around the world.

Third, the framework presented above focused on labor demand and market wages as a measure of shared prosperity and ignored other aspects of job quality. On the one hand, this position is justified because for the majority of the human population, labor is the main asset that allows them to earn income. AI applications that reduce the demand for human labor undermine the value of what is the main and, for much of the world's population, the only asset which people have. On the other hand, “quality” of labor demand, not only its volume, matters to human well-being. If an AI-induced change to production processes replaces 1 million “good” jobs with 3.5 million jobs of equal pay but starkly lower job quality (e.g.\ work that is precarious, dehumanizing, etc.), it would be difficult to argue that this was overall beneficial to workers and advanced shared prosperity.

Lastly, even though paid labor is currently the main source of income for the majority of humans, we do not want to rule out the possibility that far greater \emph{shared prosperity} and human flourishing may be possible in the future of humanity, if all labor is replaced by sufficiently intelligent AI systems and machines \cite{tegmark17, trammellk20}. This would entail a transformative change in our society and would bring about a stark set of novel challenges, including challenges in the way in which income is distributed \cite{korinekj21}. We believe that in the short and medium run, while paid labor is the main source of income for the majority, our proposed framework to develop AI for shared prosperity offers one of the most promising ways of distributing the gains from technological progress. If we did approach a situation in which novel AI systems turn all human labor into a redundant technology, we believe -- in the same spirit -- that it would be incumbent upon the developers of these systems to develop new ways of sharing the economic gains generated by their inventions that do not rely on compensating labor in order to avoid mass immiseration and advance shared prosperity globally across human society.

\section{Conclusion}
Advances in AI have the potential to produce far-reaching impacts on workers and carry the risk of exacerbating long-standing inequalities within as well as between countries. AI developers have a moral responsibility to think about the economic impact of their creations, whom they might benefit and whom they might harm. We argue that the Responsible AI community needs to make an effort to develop frameworks and heuristics for thinking rigorously about these impacts and to steer AI development choices away from applications that deepen economic inequality and instead into directions that fulfill the potential of AI to generate shared prosperity.

\begin{acks}
Klinova acknowledges the support from the Ford Foundation. Korinek acknowledges financial support from the Partnership on AI (PAI)'s Shared Prosperity Initiative (SPI). The views expressed in this work do not necessarily reflect the views of the PAI, the AI SPI, or its Members.
\end{acks}

\bibliographystyle{ACM-Reference-Format}
\bibliography{AI_SPI_references.bib}

\end{document}